\documentclass{article}

\usepackage{arxiv}
\usepackage{tabularx}
\usepackage{longtable}

\usepackage[utf8]{inputenc} 
\usepackage[T1]{fontenc}    
\usepackage{hyperref}       
\usepackage{url}            
\usepackage{booktabs}       
\usepackage{amsfonts}       
\usepackage{nicefrac}       
\usepackage{microtype}      
\usepackage{lipsum}		
\usepackage{graphicx}
\usepackage{subcaption}
\usepackage{dblfloatfix}
\usepackage{rotating}
\usepackage{xcolor,colortbl}
\usepackage{amsmath}

\usepackage{multirow}
\usepackage{enumerate} 
\usepackage{xcolor,colortbl}
\usepackage{enumitem}
\usepackage{bm}
\usepackage[linesnumbered,ruled]{algorithm2e}
\usepackage{pifont}
\usepackage{amssymb}
\newtheorem{theorem}{Theorem}

\def\x{{\boldsymbol x}}
\def\y{{\boldsymbol y}}
\def\z{{\boldsymbol z}}

\DeclareMathOperator*{\argmax}{arg\,max}

\definecolor{Gray}{gray}{0.9}
\definecolor{LightCyan}{rgb}{0.88,1,1}

\newcolumntype{a}{>{\columncolor{Gray}}c}
\newcolumntype{b}{>{\columncolor{white}}c}

\newcommand*{\belowrulesepcolor}[1]{%
  \noalign{%
    \kern-\belowrulesep 
    \begingroup 
      \color{#1}%
      \hrule height\belowrulesep 
    \endgroup 
  }%
} 
\newcommand*{\aboverulesepcolor}[1]{%
  \noalign{%
    \begingroup 
      \color{#1}%
      \hrule height\aboverulesep 
    \endgroup 
    \kern-\aboverulesep 
  }%
} 

\newtheorem{lemma}[theorem]{Lemma}

\SetCommentSty{mycommfont}

\begin{document}
%
\title{Prototypical Classifier for Robust Class-Imbalanced Learning}

\author{
    Tong Wei$^1$, \quad Jiang-Xin Shi$^1$, \quad Yu-Feng Li$^1$, \quad Min-Ling Zhang$^2$  \\
    $^1$Nanjing University, Nanjing, China \\
    $^2$Southeast University, Nanjing, China \\
    \texttt{\{weit,shijx\}@lamda.nju.edu.cn}\\
}
%
%
\maketitle              
\begin{abstract}

Deep neural networks have been shown to be very powerful methods for many supervised learning tasks. However, they can also easily overfit to training set biases, i.e., label noise and class imbalance. While both learning with noisy labels and class-imbalanced learning have received tremendous attention,
existing works mainly focus on one of these two training set biases. To fill the gap, we propose \textit{Prototypical Classifier}, which does not require fitting additional parameters given the embedding network. Unlike conventional classifiers that are biased towards head classes, Prototypical Classifier produces balanced and comparable predictions for all classes even though the training set is class-imbalanced. By leveraging this appealing property, we can easily detect noisy labels by thresholding the confidence scores predicted by Prototypical Classifier, where the threshold is dynamically adjusted through the iteration. A sample reweghting strategy is then applied to mitigate the influence of noisy labels. We test our method on CIFAR-10-LT, CIFAR-100-LT and Webvision datasets, observing that Prototypical Classifier obtains substaintial improvements compared with state of the arts.

\end{abstract}

\section{Introduction}

Deep neural networks (DNNs) have been widely used for machine learning applications. Despite of their success, it has been shown that the training of DNNs requires large-scale labeled and \textit{unbiased} data. However, in many real-world applications, training set biases are prevalent~\cite{DBLP:conf/icml/RenZYU18,weit2020tnnls,Wei_2021_RoLT,DBLP:journals/corr/abs-2108-11096}, which typically have two types: i) class-imbalanced data distribution; and ii) noisy labels. For example, in autonomous driving, the vast majority of the training data is composed of standard vehicles but models also need to recognize rarely seen classes such as
emergency vehicles or animals with very high accuracy. This will sometime lead to biased training models that do not perform well in practice. Moreover, large-scale high-quality data annotations are expensive and time-consuming to obtain. Although coarse labels are cheap and of high availability, the presence of noise will hurt the model performance. Therefore, it is desirable to develop machine learning algorithms that can accommodate not only class-imbalanced training set, but also the presence of label noise.

Both learning with noisy labels and class-imbalanced learning (a.k.a. long-tailed learning) have been studied for many years. When dealing with label noise, the most popular approach is sample selection where correctly-labeled examples are identified by capturing the training dynamics of DNNs~\cite{Li_2021_ICCV,Wu_2021_ICCV}. When dealing with class imbalance, many existing works propose to reweight examples or design unbiased loss functions by taking into account the class distribution of training set~\cite{DBLP:conf/nips/WangRH17,DBLP:conf/cvpr/CuiJLSB19,DBLP:conf/iclr/KangXRYGFK20}. However, most existing methods focus on only one of these two training set biases.

\begin{figure}[t]
    \centering
    \begin{subfigure}[b]{0.29\textwidth}
        \centering
        \includegraphics[width=\linewidth]{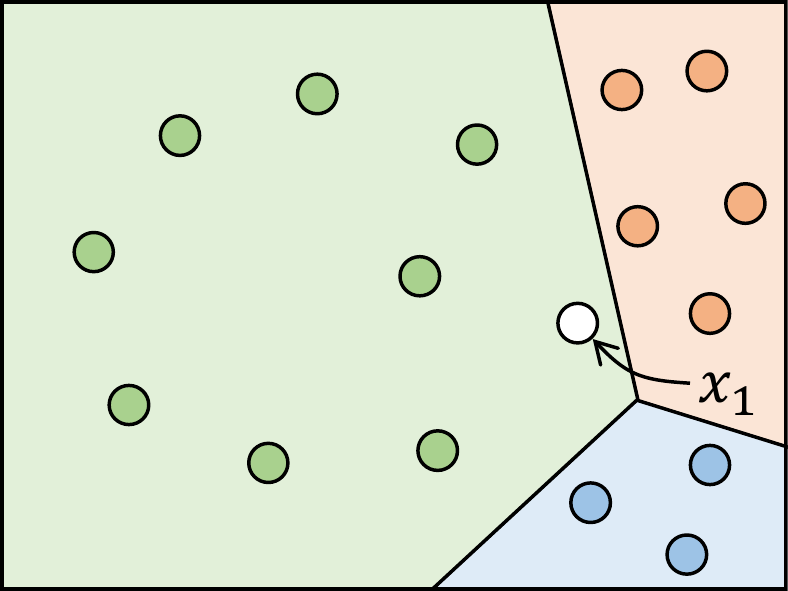}
        \caption{Normal Classifier} \label{fig:pic_erm}
    \end{subfigure}
    \hspace{0.1cm}
    \begin{subfigure}[b]{0.29\textwidth}
        \centering
        \includegraphics[width=\linewidth]{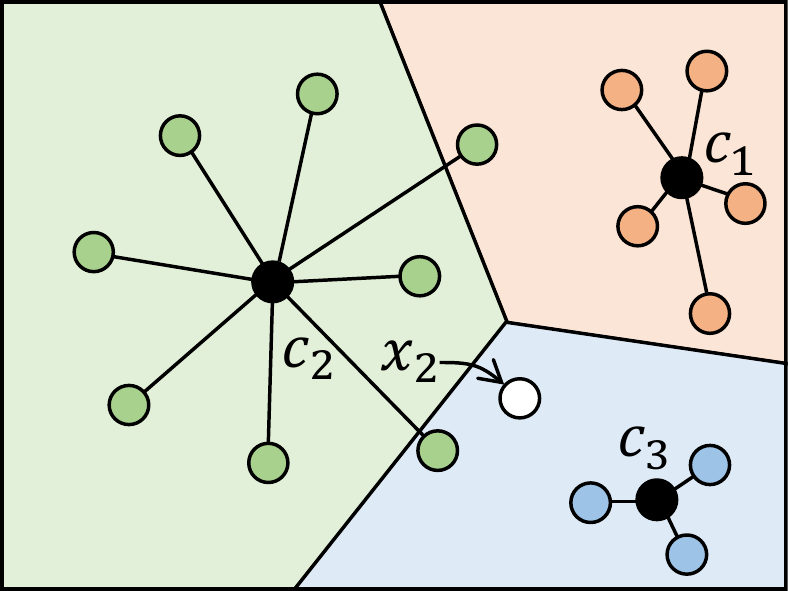}
        \caption{Prototypical $1$-NN} \label{fig:pic_ncm}
    \end{subfigure}
    \hspace{0.1cm}
    \begin{subfigure}[b]{0.29\textwidth}
        \centering
        \includegraphics[width=\linewidth]{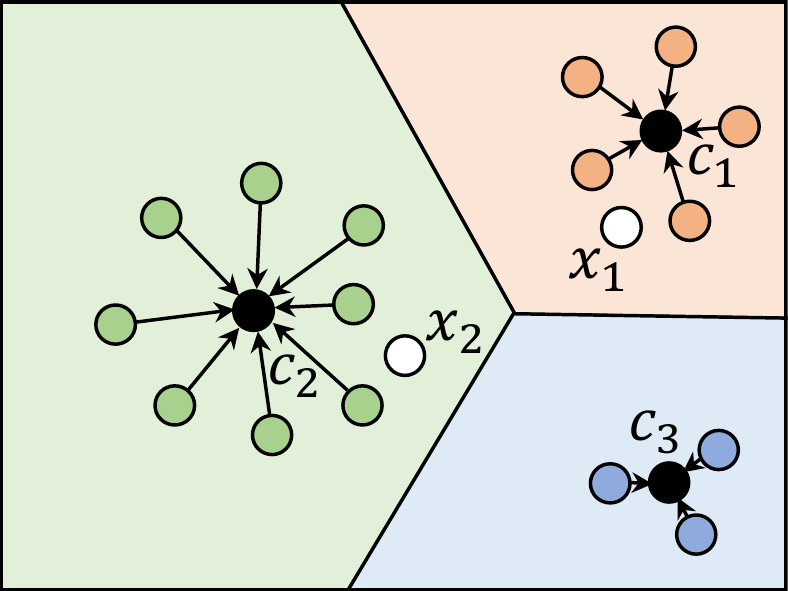}
        \caption{Prototypical Classifier} \label{fig:pic_ours}
    \end{subfigure}
    \caption{Illustration of normal classifier and Prototypical Classifier.}
\end{figure}

In this paper, we address both training set biases simultaneously. As shown in Figure~\ref{fig:pic_erm}, it is known that the classifier directly learned on class-imbalanced data is biased towards head classes~\cite{DBLP:conf/iclr/KangXRYGFK20,DBLP:conf/cvpr/ZhouCWC20bbn} which results in poor generalization on tail classes. Moreover, using sample loss/confidence produced by biased classifiers fails to detect label noise, because both clean and noisy samples of tail classes have large loss and low confidence. To solve this problem, we propose to use \textit{Prototypical Classifier} which is demonstrated to produce balanced predictions even through the training set is class-imbalanced. Our basic idea is that there exists an embedding in which examples cluster around a single prototype representation for each class. In order to do this, we learn a non-linear mapping of the input into an embedding space using a neural network and take a class's prototype to be the normalized mean vector of examples in the embedding space. Classification is then performed for an embedded test example by simply finding the nearest class prototype. Notably, Prototypical Classifier does not need additional learnable parameters given embedding of examples. Unfortunately, it is easy to observe that simply using prototypes for classification may lead to many wrong predictions for samples of head classes as shown in Figure~\ref{fig:pic_ncm}. The reason is that the representations are supposed to be modified when the classification boundaries of tail classes expand. We therefore train the neural networks to pull together embedding of examples and the prototype of their class, while pushing apart examples from prototypes of other classes. By doing this, it can avoid many mis-classifications for samples of head classes, as shown in Figure~\ref{fig:pic_ours}.
Subsequently, we find that the confidence scores produced by Prototypical Classifier is balanced and comparable across classes. By leveraging this property, we can simply detect noisy labels via thresholding where the threshold is dynamically adjusted, followed by a sample re-weighting strategy.

In summary, our key contributions of this work are:
\begin{itemize}
    \item We propose to learn from training set with mixed biases, which is practical but has been understudied;
    \item Our approach, Prototype Classifier, is simple yet powerful. It produces more balanced predictions over all classes than normal classifiers even when the training set is class-imbalanced. This property further benefits the detection of label noise.
    \item On both simulated datasets and a real-world dataset Webvision with label noise, Prototype Classifier achieves substaintial performance improvement.
\end{itemize}

\section{Related Work}


\textbf{Class-Imbalanced Learning.} Recently, many approaches have been proposed to handle class-imbalanced training set. Most extant approaches can be categorized into three types by modifying (i) the inputs to a model by re-balancing the training data~\cite{DBLP:conf/eccv/ShenLH16,DBLP:conf/cvpr/0002MZWGY19,DBLP:conf/cvpr/ZhouCWC20bbn}; (ii) the outputs of a model, for example by post-hoc adjustment of the classifier~\cite{DBLP:conf/iclr/KangXRYGFK20,tang2020tde,menon2021logitadjust}; and (iii) the internals of a model by modifying the loss function~\cite{DBLP:conf/nips/CaoWGAM19ldam,DBLP:conf/nips/ShuXY0ZXM19,jamal2020rethinking,DBLP:conf/nips/RenYSMZYL20}. Each of the above methods are intuitive, and have shown strong empirical performance. However, these methods assume the training examples are correctly-labeled, which is often difficult to obtain in real-world applications. Instead, we study a realistic problem to learn from class-imbalanced data with label noise. 

\textbf{Label Noise Detection.} Plenty of methods have been proposed to detect noisy labels~\cite{jiang2018mentornet,han2018co,li2020dividemix}. Many works adopt the small-loss trick, which treats samples with small training losses as correctly-labeled. In particular, MentorNet~\cite{jiang2018mentornet} reweights samples with small loss so that noisy samples contribute less to the loss.  Co-teaching~\cite{han2018co} trains two networks where each network selects small-loss samples in a mini-batch to train the other. DivideMix~\cite{li2020dividemix} fits a Gaussian mixture model on per-sample loss distribution to divide the training data into clean set and noisy set. In addition, AUM~\cite{pleiss2020identifying} introduces a margin statistic to identify noisy samples by measuring the average difference between the logit values for a sample's assigned class and its highest non-assigned class. The above methods only consider class-balanced training sets, thus is not directly applicable for class-imbalanced problems. Ref.~\cite{li2021mopro} observes that real-world dataset with label noise also has imbalanced number of samples per-class. Nevertheless, they only inspect a particular setup of class imbalance.

\section{Prototypical Classifier with Dynamic Threshold}

\subsection{Motivation}

\begin{figure}[htbp]
    \centering
    \begin{subfigure}[b]{0.42\textwidth}
        \centering
        \includegraphics[width=\linewidth]{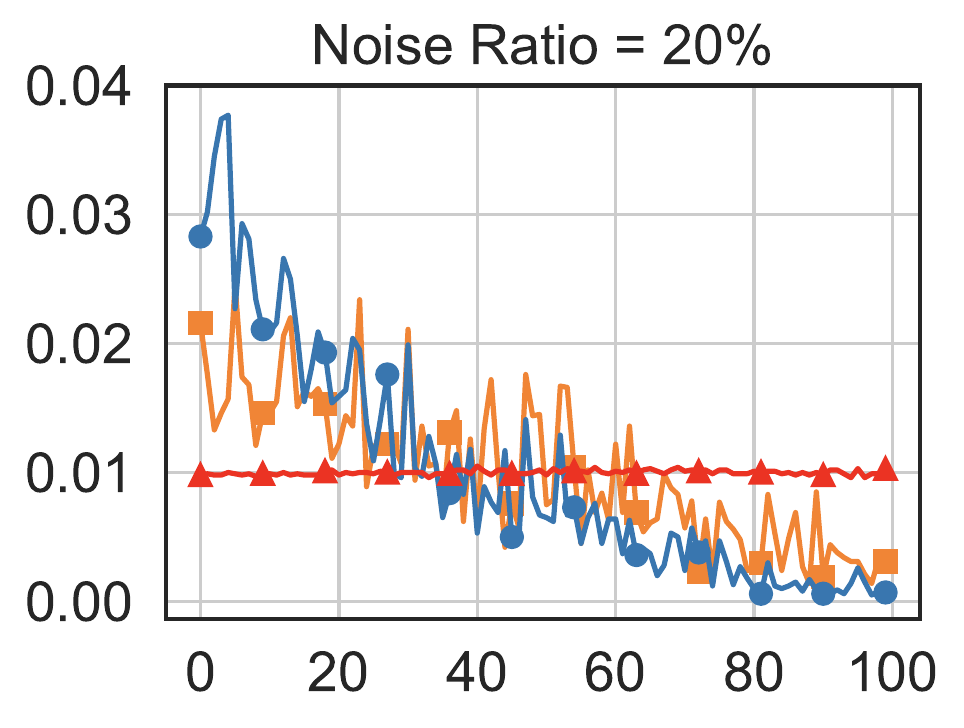}
    \end{subfigure}
    \hspace{0.3cm}
    \begin{subfigure}[b]{0.42\textwidth}
        \centering
        \includegraphics[width=\linewidth]{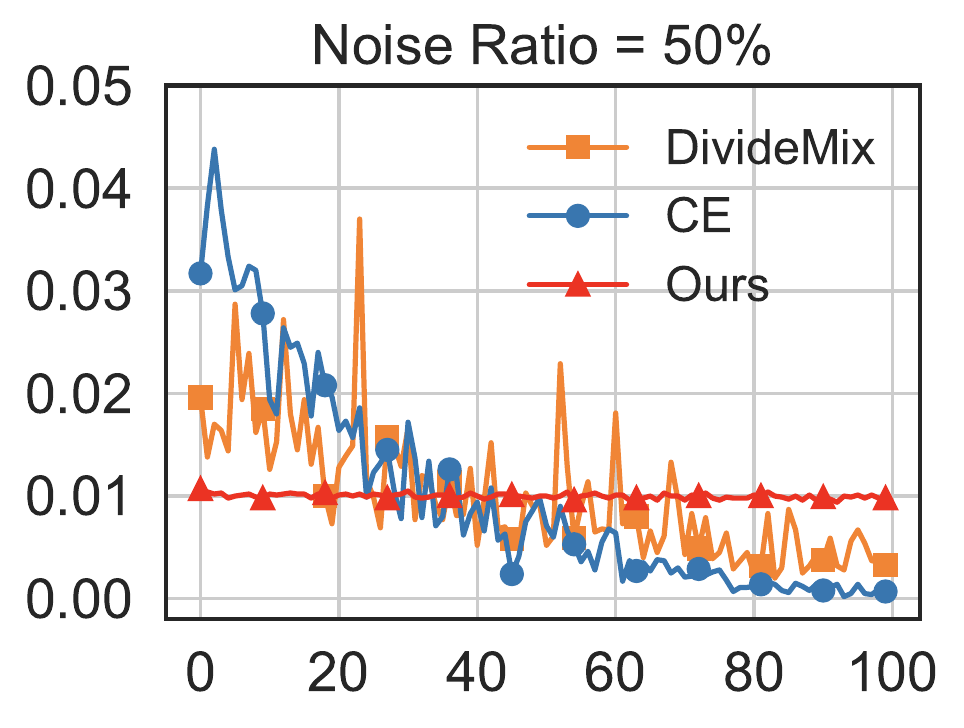}
    \end{subfigure}
    \caption{Experiment on CIFAR-100-LT. x-axis is the class labels with decreasing training samples and y-axis is the marginal likelihood $p(y)$ on the test set.}\label{fig:marginal_prob}
\end{figure}

Consider a binary classification problem with the data generating distribution $\mathbb{P}_{XY}$ being a mixture of two Gaussians. In particular, the label $Y$ is either positive (+1) or negative (-1) with equal probability (i.e., $\frac{1}{2}$). Condition on $Y = +1, \mathbb{P} (X \mid Y = +1) \sim \mathcal{N} (\mu_1, \sigma_1)$ and similarly, $\mathbb{P} (X \mid Y = -1) \sim \mathcal{N} (\mu_2, \sigma_2)$.
Without loss of generality, let $\mu_1 > \mu_2$. It is straightforward to verify that the optimal Bayes's classifier is $f(x) = sign(x - \frac{\mu_1+\mu_2}{2})$~\cite{DBLP:conf/nips/YangX20rethinking}, i.e., classify $x$ as +1 if $x > \frac{\mu_1+\mu_2}{2}$. This reminds us the nearest neighbor classifier, whose classification boundary is at the middle of two data points (i.e., balanced classification boundary). For general multi-class tasks, this motivates us to measure the distance of samples to class prototypes, which is empirically observed to produce balanced classification boundary even though the training set is class-imbalanced, as shown in Figure~\ref{fig:marginal_prob}.

In order to do this, we learn a non-linear mapping of the input into an embedding space using a neural network $f_{\theta}$ parameterized by $\theta$ using training set $\mathcal{D} = \{(\x_i, y_i)\}_{i=1}^N$. The class prototype is taken as the normalized mean vector of the embedded examples belonging to its class. For example, the prototype for class $k \in \{1,\dots, K\}$ is computed as:
\begin{equation}\label{equ:compute-prototype}
\boldsymbol{c}_{k} = \operatorname{Normalize}\bigg( \frac{1}{|\mathcal{D}_k|} \sum_{ i \in \mathcal{D}_k } f_\theta (\x_i) \bigg), \mathcal{D}_k = \left\{ i \mid y_{i}=k \right\}.
\end{equation}
Prototypical Classifier produces a distribution over classes for sample $\x$ based on a softmax over distances to the prototypes in the embedding space. In particular, when use cosine similarity as distance measure, we have:
\begin{equation}\label{equ:prototype-classifier}
\mathbb{P}_{\theta}( Y=k \mid \x)=\frac{\exp \left(f_{\theta}(\x)^{\top} \mathbf{c}_{k}\right)}{\sum_{k^{\prime}} \exp \left(f_{\theta}(\x)^{\top} \mathbf{c}_{k^{\prime}}\right)}.
\end{equation}
Learning proceeds by minimizing the negative log-probability $J(\theta)=-\log \mathbb{P}_{\theta}(Y=k \mid \mathbf{x})$ of the true class label $k$ via SGD.
Notably, the model in Equation~\eqref{equ:prototype-classifier} is equivalent to a linear model with a particular parameterization~\cite{DBLP:journals/pami/MensinkVPC13}. To see this, expand the term in the exponent:
\begin{equation}
\mathbf{c}_{k}^{\top} f_{\theta}(\x) = \mathbf{w}_{k}^{\top} f_{\theta}(\x)+b_{k}, \text { where } \mathbf{w}_{k}= \mathbf{c}_{k} \text { and } b_{k}=0.
\end{equation}
Our results indicate that Prototypical Classifier is effective despite the equivalence to a linear model. We hypothesize this is because all of the required non-linearity can be learned within the embedding function~\cite{DBLP:conf/nips/SnellSZ17}. Indeed, this is the approach that modern neural network classification systems currently use.

\subsection{Dynamic Thresholding for Label Noise Detection}
\begin{figure}[htbp]
    \centering
    \begin{subfigure}[b]{0.46\textwidth}
        \centering
        \includegraphics[width=\linewidth]{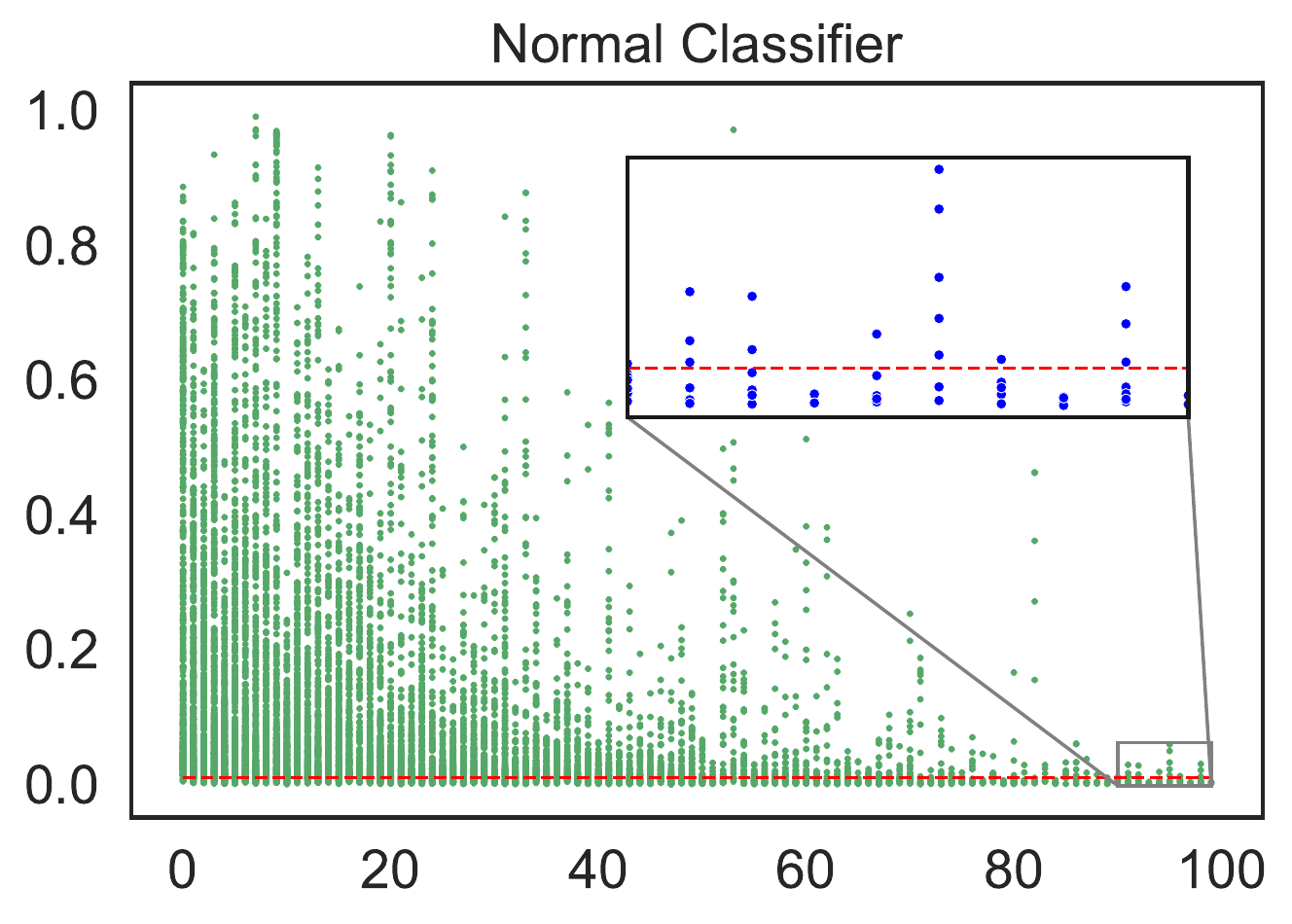}
    \end{subfigure}
    \hspace{0.1cm}
    \begin{subfigure}[b]{0.46\textwidth}
        \centering
        \includegraphics[width=\linewidth]{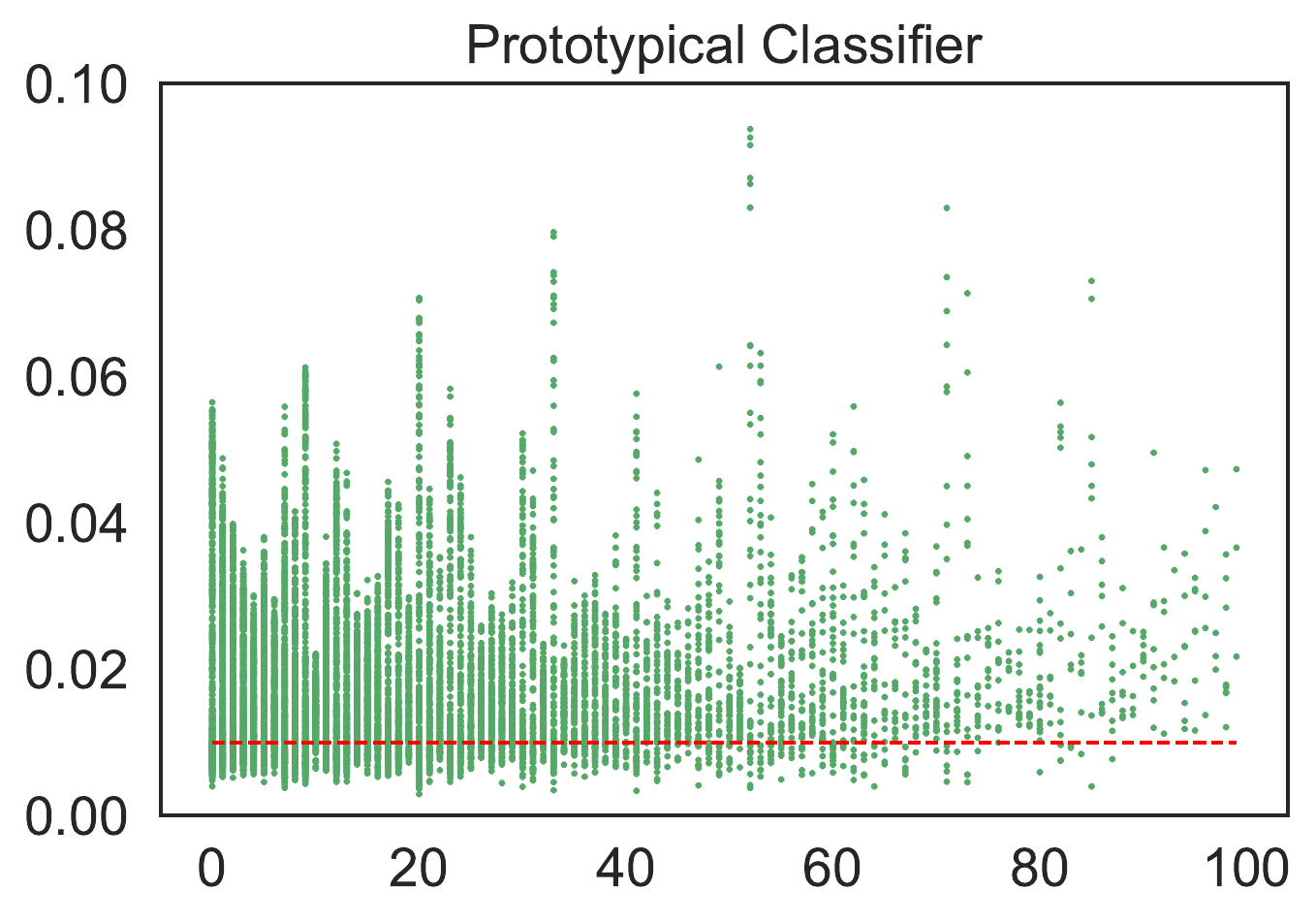}
    \end{subfigure}
    \caption{Experiment on CIFAR-100-LT. x-axis is the class labels with decreasing training samples and y-axis is the confidence scores of classifiers on training set.}\label{fig:predicted_scores}
\end{figure}

\begin{figure}[h]
    \centering
    \begin{subfigure}[b]{0.4\textwidth}
        \centering
        \includegraphics[width=\linewidth]{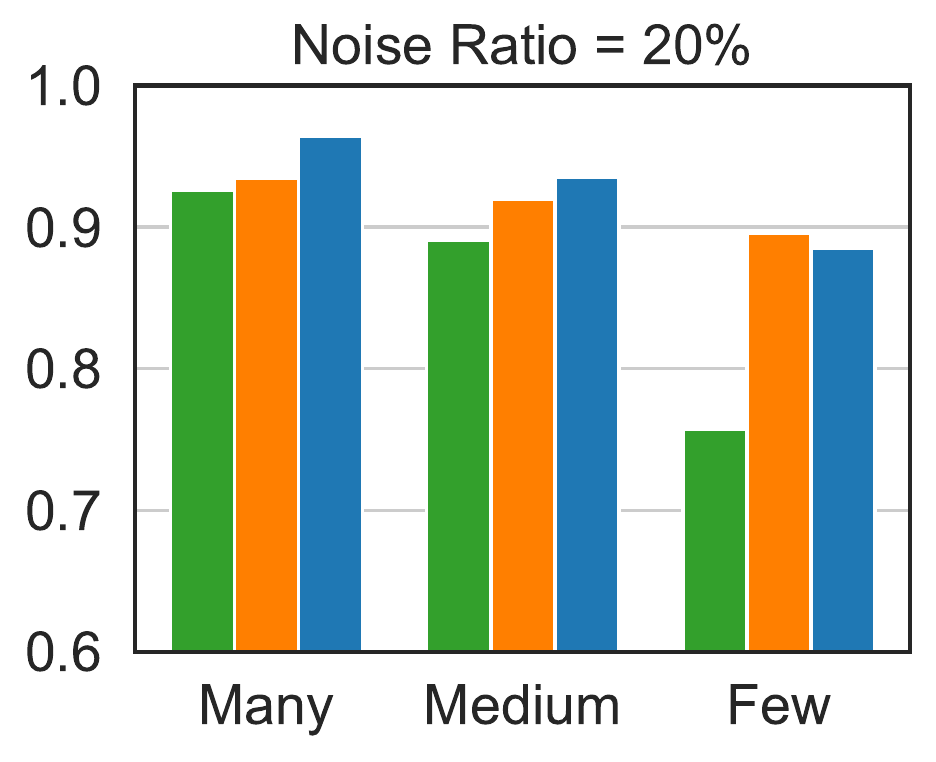}
    \end{subfigure}
    \hspace{0.1cm}
    \begin{subfigure}[b]{0.4\textwidth}
        \centering
        \includegraphics[width=\linewidth]{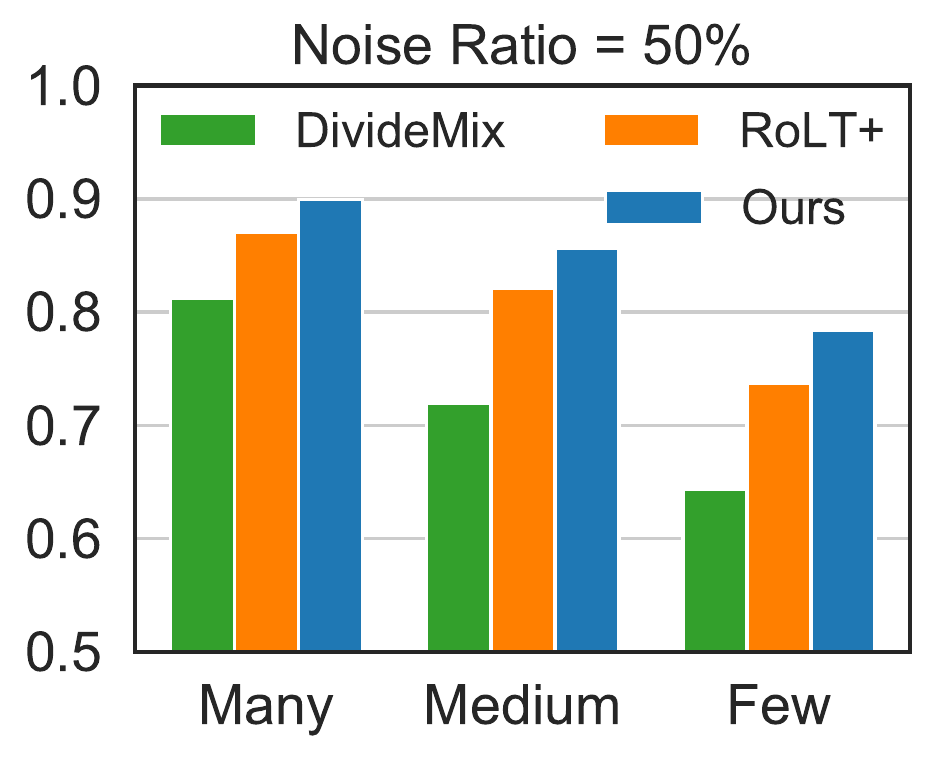}
    \end{subfigure}
    \caption{Experiment on CIFAR-100-LT. We show the F$_1$-score of clean examples selection module for many, medium and few classes.}\label{fig:f1}
\end{figure}

However, the existence of label noise may hurt the representation learning of the network. To tackle this issue, it is a common practice to correct noisy labels.
Let $\hat{\y} = [\hat{y}_1, \cdots, \hat{y}_K] = \mathbb{P}_{\theta}( Y \mid \x)$ be the prediction of Prototypical Classifier, the labels are refined as stated by the following rule:
\begin{equation}\label{equ:soft-label}
\tilde{y} = \left\{\begin{array}{ll}
y_{i} & \text { if } \hat{y}_{y_i} > \tau_t \\
\argmax_j \hat{y}_{j} & \text { otherwise. }
\end{array}\right.
\end{equation}
In words, we deem samples as clean if the confidence scores on their original labels is greater than a threshold $\tau_t$. It is notably that using normal classifiers cannot achieve this goal due to its biased predictions, while predictions of Prototypical Classifier are balanced and comparable. We illustrate this finding in Figure~\ref{fig:predicted_scores}.
We then need to construct $\tau_t$. Intuitively, with the increase of the optimization iteration $t$, the predictive confidence also increases in general, so that $\tau_t$ is also required to increase.
Mathematically, we set the dynamic threshold $\tau_t$ as an increasing function of $t$, which is given by:
\begin{equation}\label{equ:tau}
    \tau_t = \gamma^{t} \tau_0.
\end{equation}
Here, $\tau_0$ is the initial threshold and $\gamma$ is set to $1.005$ in our experiments. We provide more analysis about $\tau_t$ in supplementary materials. Lemma~\ref{lemma} summarizes the performance bound of the label noise detection method.
\begin{lemma}\label{lemma}
With probability at least $p$, the F$_1$-score of detecting noisy labels in $\mathcal{D}_j$ by thresholding the predictive scores of Prototypical Classifier is at least $1-\frac{e^{-v} \max \left(N^{-}, N^{+}\right)+\alpha}{N^{-}}$ when the noise ratio is known, where $p=\int_{-1}^{\mu^{t r u e}-\mu^{f a l s e}-\Delta} f(t) d t$, $f(t)$ is the probability density function of the difference of two independent beta-distributed random variables $\beta_1 - \beta_2$, where $\beta_{1} \sim \operatorname{Beta}\left(N^{-}, 1\right)$, $\beta_{2} \sim \operatorname{Beta}\left(\alpha+1, N^{+}-\alpha\right)$. 
\end{lemma}
Lemma~\ref{lemma} shows that the performance of noise detection depends on the intraclass concentration of clean samples in the embedding space (denoted by $\frac{\Delta^2}{v}$), which is optimized by the prototypical contrastive loss defined in Equation~\eqref{equ:wpc}. The theory was first shown in Ref.~\cite{zhu2021good}. We further justify the effectiveness of our method in Figure~\ref{fig:f1}, which produces high F$_1$-score for both head and tail classes. 

\subsection{Example Reweighting}
In standard training, we aim to minimize the expected loss
for the training set, where each input example is weighted equally. Here we aim to learn a reweighting of the inputs to cope with hard mislabeled samples whose labels are not correctly refined, where we minimize a weighted loss:
\begin{equation}\label{equ:wpc}
\mathcal{L}_{\text {pc }}= \frac{-1}{ \sum_{i=1}^N w_i } \sum_{i=1}^{N} w_i \log \frac{\exp \left(f_{\theta}(\x) \cdot \boldsymbol{c}_{y_{i}} / \tau\right)}{\sum_{k=1}^{K} \exp \left(f_{\theta}(\x) \cdot \boldsymbol{c}_{k} / \tau\right)}.
\end{equation}
With a slight abuse of the notation, we re-define $w_i$ to be the weight for the $i$-th example and $\tau$ is a temperature parameter. We expect the weights can reflect the likelihood of examples being correctly-labeled. In that regard, we devise a weighted version for computing prototypes as:
\begin{equation}\label{equ:weighted-compute-prototype}
\boldsymbol{c}_{k} = \operatorname{Normalize}\bigg( \frac{1}{\sum_{i \in \mathcal{D}_k} w_i} \sum_{ i \in \mathcal{D}_k } w_i f_\theta (\x_i) \bigg), \mathcal{D}_k = \left\{ i \mid y_{i}=k \right\}.
\end{equation}
Recall that, one appealing property of Prototypical Classifier is balanced predictions across all classes, as opposite to biased normal classifiers.
We therefore simply set examples weights as the predicted score of Prototypical Classifier on the training label, i.e., for the $i$-th example, we set $w_i = \mathbb{P}_{\theta}(Y=y_i \mid \x_i)$ where $y_i$ is the training label of $\x_i$. For samples whose labels are rectified, we update their weights by $w' = \frac{\tau_t - w}{2}$ to reflect the uncertainty. The modified example weights are always positive since the label is refined if and only if $w = \mathbb{P}(Y = y_i \mid \x_i) \leq \tau_t$.
The optimization of $\mathcal{L}_{\text {pc}}$ is realized by contrastive learning, which has been demonstrated effective in learning representations~\cite{DBLP:conf/iclr/0001ZXH21}.
Observing that the presence of label noise may have negative effect on representation learning, we train networks to optimize the unsupervised contrastive loss, which does not use the biased training labels. The basic idea of unsupervised contrastive learning is to pull together two embeddings of the same example, while pushing apart from other examples. Formally, let $\z_i = f_{\theta}(\x_i)$ and $\z_i^{\prime}$ be the embedding of augmented version of $\x_i$, the unsupervised contrastive loss is computed as:
\begin{equation}
\mathcal{L}_{\text {cc }}^{i}=-\log \frac{\exp \left(\boldsymbol{z}_{i} \cdot \boldsymbol{z}_{i}^{\prime} / \tau\right)}{\sum_{b=0}^{B} \exp \left(\z_i \cdot \boldsymbol{z}_{b}^{\prime} / \tau\right)},
\end{equation}
where $\tau$ is a scalar temperature parameter and $B$ is mini-batch size.

Given the above definitions and denoting $\mathcal{L}^{\mathrm{ce}}$ as conventional cross-entropy loss, the overall training objective is written as:
\begin{equation}
\mathcal{L}=\mathcal{L}^{\mathrm{ce}}+\lambda_{1} \mathcal{L}^{\mathrm{cc}}+\lambda_{2} \mathcal{L}^{\mathrm{pc}},
\label{eq:total_loss}
\end{equation}
where hyperparameters $\lambda_{1}$ and $\lambda_{2}$ are trade-off parameters. 
We adopt DNNs as feature extractor and a linear layer as projector to generate latent feature representation $\boldsymbol{z}_i$. Another linear layer following the feature extractor is used as classifier. When minimizing $\mathcal{L}_{\mathrm{pc}}$, we apply mixup~\cite{zhang2017mixup} to improve the generalization which has been shown to be effective for learning with noisy labels~\cite{Wu_2021_ICCV}.

\section{Experiments}
We perform experiments on CIFAR-10 and CIFAR-100 datasets by controlling label noise ratio and imbalance factor of the training set.
Additionally, we perform experiments on a commonly used dataset Webvision with real-world label noise. 

\subsection{Results on Simulated Datasets}

\textbf{Class-Imbalanced Dataset Generation.}
Formally, for a dataset with $K$ classes and $N$ training examples for each class, by assuming the imbalance factor is $\rho$, the number of examples for the $k$-th class is set to $N_k={N}/{\rho^{\frac{k-1}{K-1}}}$.

\textbf{Label Noise Injection.}
Let $Y$ denote the variable for the clean label, $\bar{Y}$ the noisy label, and $X$ the instance/feature, the transition matrix $T(X = x)$ is defined as $T_{ij}(X) = \mathbb{P}(\bar{Y}=j \mid Y=i, X=x)$. In this work, we follow the setup in RoLT+~\cite{Wei_2021_RoLT} by setting $T(X = x)$ according to the estimated class priors $\mathbb{P}(y)$, e.g., the empirical class frequencies in the training dataset. Formally, given the noise proportion $\gamma \in [0,1]$, we define:
\begin{equation}\label{equ:transition-matrix}
    T_{ij}(X) = \mathbb{P}(\bar{Y}=j \mid Y=i, X=x) = \left\{\begin{array}{ll}
1 - \gamma  & i = j \\
\frac{N_j}{N - N_i} \gamma & \text { otherwise. }
\end{array}\right.
\end{equation}
Here, $N$ is the size of training set and $N_j$ is frequency of class $j$. 

\setlength{\tabcolsep}{5pt}
\begin{table}[h]
\centering
\begin{tabular}{ l c | c c c | c c c }
\toprule
\multicolumn{2}{ l| }{Noise Ratio} & \multicolumn{3}{ c| }{0.2} & \multicolumn{3}{ c }{0.5}\\
\hline
\multicolumn{2}{ l| }{Imbalance Factor} & 10 & 50 & 100 & 10 & 50 & 100 \\
\hline
\multirow{2}{*}{(1) CE}  & Best  & 77.86 & 64.38 & 61.79  & 60.72 & 46.50 & 38.43 \\
 & Last  & 74.00 & 61.38 & 55.69  & 44.29 & 32.69 & 27.78  \\
\hline
\multirow{2}{*}{(2) LDAM}  & Best  & 83.48 & 72.01 & 66.41  & 63.57 & 38.92 & 34.08 \\
 & Last  & 82.91 & 71.23 & 66.22  & 62.13 & 37.97 & 32.56  \\
\hline
\multirow{2}{*}{(3) LDAM-DRW}  & Best  & 84.98 & 76.77 & 73.24  & 69.53 & 49.90 & 42.60 \\
 & Last  & 84.71 & 75.98 & 72.46  & 68.76 & 47.71 & 40.47  \\
\hline
\multirow{2}{*}{(4) DivideMix$^*$}  & Best  & 88.79 & 75.34 & 66.90  & 87.54 & 67.92 & 61.81 \\
 & Last  & 88.10 & 73.48 & 63.76  & 86.88 & 65.22 & 59.65  \\
\hline
\multirow{2}{*}{(5) RoLT+$^*$}  & Best  & 87.95 & 77.26 & 72.31  &\bf 88.17 &\bf 75.11 & 64.42 \\
 & Last  & 87.54 & 75.90 & 69.12  &\bf 87.45 &\bf 73.92 & 61.15\\
\hline
\multirow{2}{*}{(6) Prototypical Classifier}  & Best  &\bf 90.92 &\bf 84.12 &\bf 79.54 & 84.04 & 71.44 &\bf 66.33 \\
 & Last  &\bf 90.81 &\bf 83.71 &\bf 78.34 & 83.51 & 71.44 &\bf 64.69 \\
\bottomrule
\end{tabular}
\caption{ Test accuracy (\%) on CIFAR-10. $^*$ denotes ensemble models. }\label{tab:cifar10}
\end{table}

\begin{table}[h]
\centering
\begin{tabular}{ l c | c c c | c c c }
\toprule
\multicolumn{2}{ l| }{Noise Ratio} & \multicolumn{3}{ c| }{0.2} & \multicolumn{3}{ c }{0.5}\\
\hline
\multicolumn{2}{ l| }{Imbalance Factor} & 10 & 50 & 100 & 10 & 50 & 100 \\
\hline
\multirow{2}{*}{(1) CE}  & Best  & 45.97 & 33.41 & 29.85  & 28.70 & 18.49 & 16.24 \\
 & Last  & 45.75 & 33.12 & 29.58  & 23.70 & 16.56 & 14.19  \\
\hline
\multirow{2}{*}{(2) LDAM}  & Best  & 47.30 & 35.70 & 32.67  & 27.86 & 17.62 & 15.68 \\
 & Last  & 47.12 & 35.50 & 32.60  & 24.20 & 17.50 & 14.73  \\
\hline
\multirow{2}{*}{(3) LDAM-DRW}  & Best  & 47.85 & 36.29 & 33.38  & 27.86 & 17.91 & 15.68  \\
 & Last  & 47.68 & 36.01 & 32.99  & 24.45 & 17.81 & 15.07  \\
\hline
\multirow{2}{*}{(4) DivideMix$^{*}$}  & Best  & 63.79 & 49.64 & 43.91  & 49.35 & 36.52 & 31.82 \\
 & Last  & 63.17 & 48.37 & 42.59  & 48.87 & 35.72 & 31.05 \\
\hline
\multirow{2}{*}{(5) RoLT+$^{*}$}  & Best  & 64.22 & 51.01 & 45.35  & 53.31 & 39.78 & 35.29 \\
 & Last  & 63.31 & 49.40 & 43.16  & 52.44 & 39.27 & 34.43 \\
\hline
\multirow{2}{*}{(6) Prototypical Classifier}  & Best &\bf 65.23 &\bf 51.73 &\bf 47.38 &\bf 57.65 &\bf 42.51 &\bf 38.42\\
 & Last  &\bf 65.14 &\bf 51.46 &\bf 47.12 &\bf 57.65 &\bf 42.51 &\bf 38.36 \\
\bottomrule
\end{tabular}
\caption{ Test accuracy (\%) on CIFAR-100. $^*$ denotes ensemble models.}\label{tab:cifar100}
\end{table}

\textbf{Result.} We train the PreAct ResNet-18 network using SGD optimizer with momentum 0.9 for all methods. We set $\lambda_1=1$ and $\lambda_2=5$. We use $\tau_0 = 0.1$ for CIFAR-10 and $\tau=0.01$ for CIFAR-100. Table~\ref{tab:cifar10} and Table~\ref{tab:cifar100} respectively summarize the results for CIFAR-10 and CIFAR-100 datasets. We compare our methods with several commonly used baselines for long-tailed learning (1-3) and learning with noisy labels (4-5). As shown in the results, previous methods dreadfully degrade their performance as the noise ratio and imbalance factor increase, while our methods retain robust performance. In particular, compared with CE, Prototypical Classifier improves the test accuracy by 9\% on average. It can be observed that the improvement becomes more significant when the noise ratio is high, benefiting from proposed noise detection method.

\begin{figure}[htbp]
    \centering
    \begin{subfigure}[b]{0.4\textwidth}
        \centering
        \includegraphics[width=\linewidth]{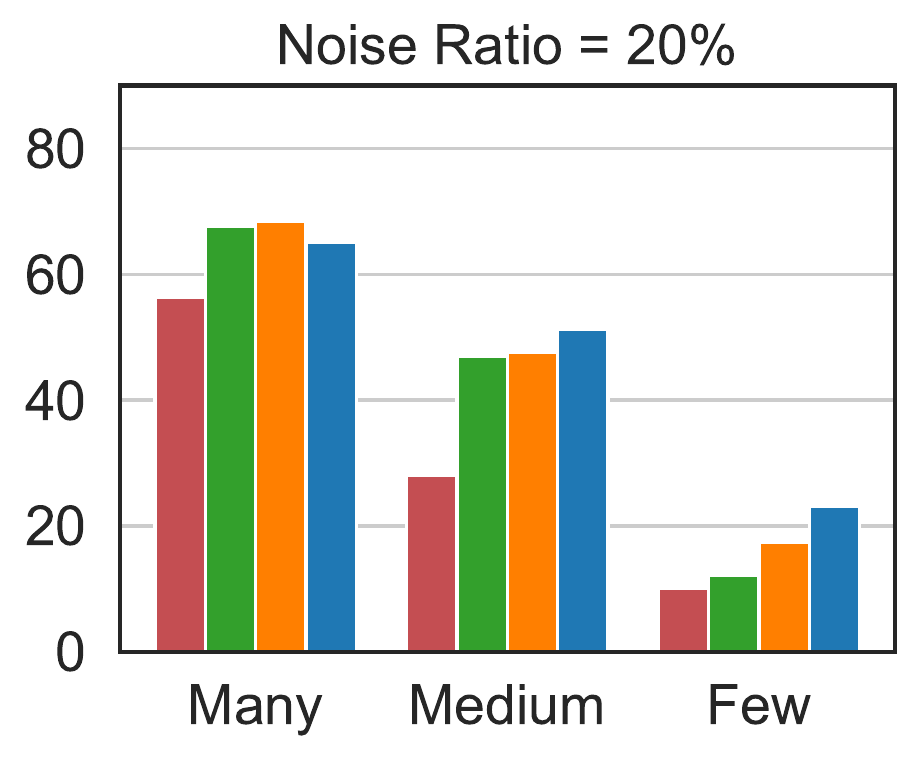}
    \end{subfigure}
    \hspace{0.1cm}
    \begin{subfigure}[b]{0.4\textwidth}
        \centering
        \includegraphics[width=\linewidth]{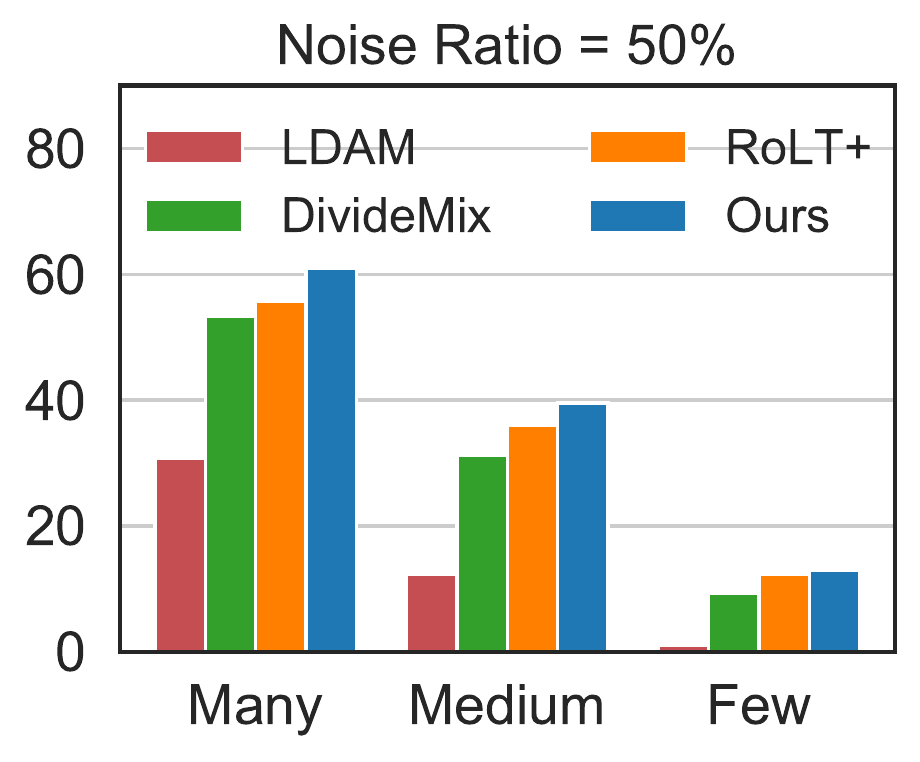}
    \end{subfigure}
    \caption{Experiment on CIFAR-100-LT. We show the accuracy for many (\#inst >100), medium (\#inst $\in [20, 100]$) and few (\#inst < 20) classes.}\label{fig:acc}
\end{figure}

As DivideMix~\cite{li2020dividemix} and RoLT+~\cite{Wei_2021_RoLT} are two strong baselines in this task, (4) and (5) obtain much higher performance than (1-3), particularly when noise ratio is high. Although (4) and (5) use an ensemble of two networks, our method (6) outperforms them in most cases. On CIFAR-100, Prototypical Classifier achieves the best results among all the approaches and outperforms others by a large margin for both head and tail classes in Figure~\ref{fig:acc}.

\subsection{Results on Real-World Dataset}

We test the performance of our method on a real-world dataset. WebVision~\cite{webvision} contains 2.4 million images collected from Flickr and Google with real noisy and class-imbalanced data. Following previous literature, we train on a subset, mini WebVision, which contains the first 50 classes. In Table~\ref{exp:webvision}, we report results comparing against state-of-the-art approaches, including MentorNet~\cite{jiang2018mentornet}, Co-teaching~\cite{han2018co}, ELR~\cite{liu2020early}, HAR~\cite{cao2021heteroskedastic}, and DivideMix~\cite{li2020dividemix}. We use InceptionResNet-v2 for all methods. We set $\tau_0 = 0.05$, $\lambda_1=1$ and $\lambda_2=2$ in all experiments.

To further uncover the advantages of our method, we run experiments by controlling the imbalance factor of Webvision dataset. The test accuracy is reported in the Table~\ref{tab:webvision-control}. From the results, we can see that the superiority of our method is more significant as the imbalance factor increases.

\begin{table}[!h]
\small
\centering
\begin{tabular}{l|c|c|c|c|c|c|c}
\toprule
& & MentorNet & Co-teaching & ELR & HAR & DivideMix$^*$ & \textbf{Ours} \\
\hline
\multirow{2}{*}{Webvision} & top1 & 63.00 & 63.58 & 76.26 & 75.5 & \textbf{77.32} & \textbf{77.32} \\
                           & top5 & 81.40 & 85.20 & 91.26 & 90.7 & 91.64 & \textbf{92.60}\\
\hline
\multirow{2}{*}{ImageNet} & top1  & 57.80 & 61.48 & 68.71 & 70.3 & \textbf{75.20} & 75.12\\
                           & top5 & 79.92 & 84.70 & 87.84 & 90.0 & 90.84 & \textbf{91.92}\\
\bottomrule
\end{tabular}
\caption{Accuracy (\%) on WebVision and ImageNet. $^*$ denotes ensemble models.}\label{exp:webvision}
\end{table}

\begin{table}[h]
    \centering
    \begin{tabular}{llcccc}
    \toprule
      Imbalance Factor   & Method & \multicolumn{2}{ c }{Webvision}  &  \multicolumn{2}{ c }{ImageNet} \\
      \cmidrule(lr){3-4}  \cmidrule(lr){5-6}
         & & top-1 & top-5 & top-1 & top-5\\
    \midrule
    \multirow[t]{4}{*}{ $\rho=50$ }    & DivideMix$^*$ & 64.56 & 83.56	& 62.68	& 85.24 \\
                                    & Prototypical Classifier & \bf 68.00 & \bf 88.44 & \bf 65.00 & \bf 86.32\\
    \hline
    \multirow[t]{4}{*}{ $\rho=100$ }   & DivideMix$^*$ & 55.76 & 73.48 & 53.92 & 74.00 \\
                                    & Prototypical Classifier & \bf 62.12 & \bf 85.88 & \bf 59.60 & \bf 84.20\\
    \bottomrule
    \end{tabular}
    \caption{ Top-1 (top-5) accuracy on Webvision. $^*$ denotes ensemble models.}
    \label{tab:webvision-control}
\end{table}

\subsection{Ablation Studies}
We examine the effectiveness of the each module of our method by removing it and comparing its performance with the full framework. The results are reported in Table~\ref{tab:ablations}. Generally, it is easy to see that removing any part of the method significantly drops the performance or even fails in some cases. The performance of re-weighting and dynamic threshold shows their great effectiveness for dealing with label noise. Though we do not use the normal classifier trained via $\mathcal{L}_{ce}$, it is observed to help improve the representation learning. We have a similar observation for the unsupervised contrastive loss $\mathcal{L}_{ce}$. The strong augmentation method AugMix~\cite{hendrycks2020augmix} also provides substaintial improvement.

Additionally, we also test our method on class-balanced training sets with label noise in Table~\ref{tab:cifar_balanced}. Prototypical Classifier outperforms other methods in most cases, even though both DivideMix and RoLT+ uses an ensemble of two networks, which shows the generality of Prototypical Classifier.

\begin{table}[htbp]
    \centering
    \begin{tabular}{cc|c|c}
    \toprule
      Method   &  & CIFAR-10  &  CIFAR-100 \\
    \hline
    \multirow{2}{*}{ w/o re-weighting }     & Best & 61.69 ($\begingroup\color{red}\blacktriangledown\endgroup$4.64) & - \\
                                            & Last & 58.57 ($\begingroup\color{red}\blacktriangledown\endgroup$6.12) & - \\
    \hline
    \multirow{2}{*}{ w/o dynamic threshold }   & Best & 63.85 ($\begingroup\color{red}\blacktriangledown\endgroup$2.48) &  39.04 ($\begingroup\color{green}\blacktriangle\endgroup$0.62)\\
                                    & Last & 56.01 ($\begingroup\color{red}\blacktriangledown\endgroup$8.68) & 38.67 ($\begingroup\color{green}\blacktriangle\endgroup$0.25) \\
    \hline
    \multirow{2}{*}{ w/o mixup }   & Best & 52.79 ($\begingroup\color{red}\blacktriangledown\endgroup$13.54) & 33.09 ($\begingroup\color{red}\blacktriangledown\endgroup$5.33) \\
                                    & Last & 51.43 ($\begingroup\color{red}\blacktriangledown\endgroup$13.26) & 32.57 ($\begingroup\color{red}\blacktriangledown\endgroup$5.79)\\
    \hline
    \multirow{2}{*}{ w/o AugMix }   & Best & 62.51 ($\begingroup\color{red}\blacktriangledown\endgroup$3.82) & 36.11 ($\begingroup\color{red}\blacktriangledown\endgroup$2.31)\\
                                    & Last & 55.21 ($\begingroup\color{red}\blacktriangledown\endgroup$9.48) & 35.68 ($\begingroup\color{red}\blacktriangledown\endgroup$2.68)\\
    \hline
    \multirow{2}{*}{ w/o $\mathcal{L}_{\mathrm{cc}}$ }   & Best & 55.34 ($\begingroup\color{red}\blacktriangledown\endgroup$9.35) & 32.65 ($\begingroup\color{red}\blacktriangledown\endgroup$5.71)\\
                                    & Last & 53.17 ($\begingroup\color{red}\blacktriangledown\endgroup$11.52) & 32.39($\begingroup\color{red}\blacktriangledown\endgroup$5.97) \\
    \hline
    \multirow{2}{*}{ w/o $\mathcal{L}_{\mathrm{ce}}$ }   & Best & 57.61 ($\begingroup\color{red}\blacktriangledown\endgroup$7.08) & 35.25 ($\begingroup\color{red}\blacktriangledown\endgroup$3.11)\\
                                    & Last & 53.24 ($\begingroup\color{red}\blacktriangledown\endgroup$11.45) & 35.02 ($\begingroup\color{red}\blacktriangledown\endgroup$3.34)\\
    \bottomrule
    \end{tabular}
    \caption{ Ablation studies. $\rho=0.5$ and $\gamma=100$. $\begingroup\color{red}\blacktriangledown\endgroup$ ($\begingroup\color{green}\blacktriangle\endgroup$) indicate performance loss (gain) compared with Prototypical Classifier.}\label{tab:ablations}
    \label{tab:ablation}
\end{table}

\setlength{\tabcolsep}{6pt}
\begin{table}[htbp]
\small
\centering
\begin{tabular}{ l c | c | c | c | c }
\toprule
 &  & \multicolumn{2}{ c| }{CIFAR-10} & \multicolumn{2}{ c }{CIFAR-100} \\
\hline
\multicolumn{2}{ l| }{Noise Ratio} & 0.2 & 0.5  & 0.2 & 0.5  \\
\hline
\multirow{2}{*}{DivideMix$^*$}  & Best  & 92.79 &\bf 95.03  & 77.25 & 73.84 \\
 & Last  & 92.41 &\bf 94.63  & 77.03 & 73.42 \\
\hline
\multirow{2}{*}{RoLT+$^*$}  & Best  & 92.46 & 94.59  & 78.60 & 74.11 \\
 & Last  & 92.01 & 94.41  & 78.14 & 73.35 \\
\hline
\multirow{2}{*}{Prototypical Classifier}  & Best  &\bf 95.93 & 92.55  &\bf 79.41 &\bf 75.50 \\
 & Last  &\bf 95.80 & 92.40  &\bf 79.41 &\bf 75.10 \\
\bottomrule
\end{tabular}
\caption{ Accuracy (\%) on class-balanced datasets. $^*$ denotes ensemble models.}\label{tab:cifar_balanced}
\end{table}

\section{Conclusion}
We propose Prototypical Classifier for learning with training set biases. Prototypical Classifier is shown to produce balanced predictions for all classes even when learned on class-imbalanced training set. This appealing property provides a way of detecting label noise by thresholding the predicted scores of examples. Experiments demonstrate the superiority of the proposed method. We believe Prototypical Classifier can motivate solutions to more problems with class-imbalanced training sets, for instance semi-supervised learning and self-supervised learning.

\bibliography{pakdd22_references}
\bibliographystyle{unsrt}

\appendix
\section{Ablations on Dynamic Threshold}
\begin{figure}[h]
    \centering
    \begin{subfigure}[b]{0.45\textwidth}
    \includegraphics[width=1\linewidth]{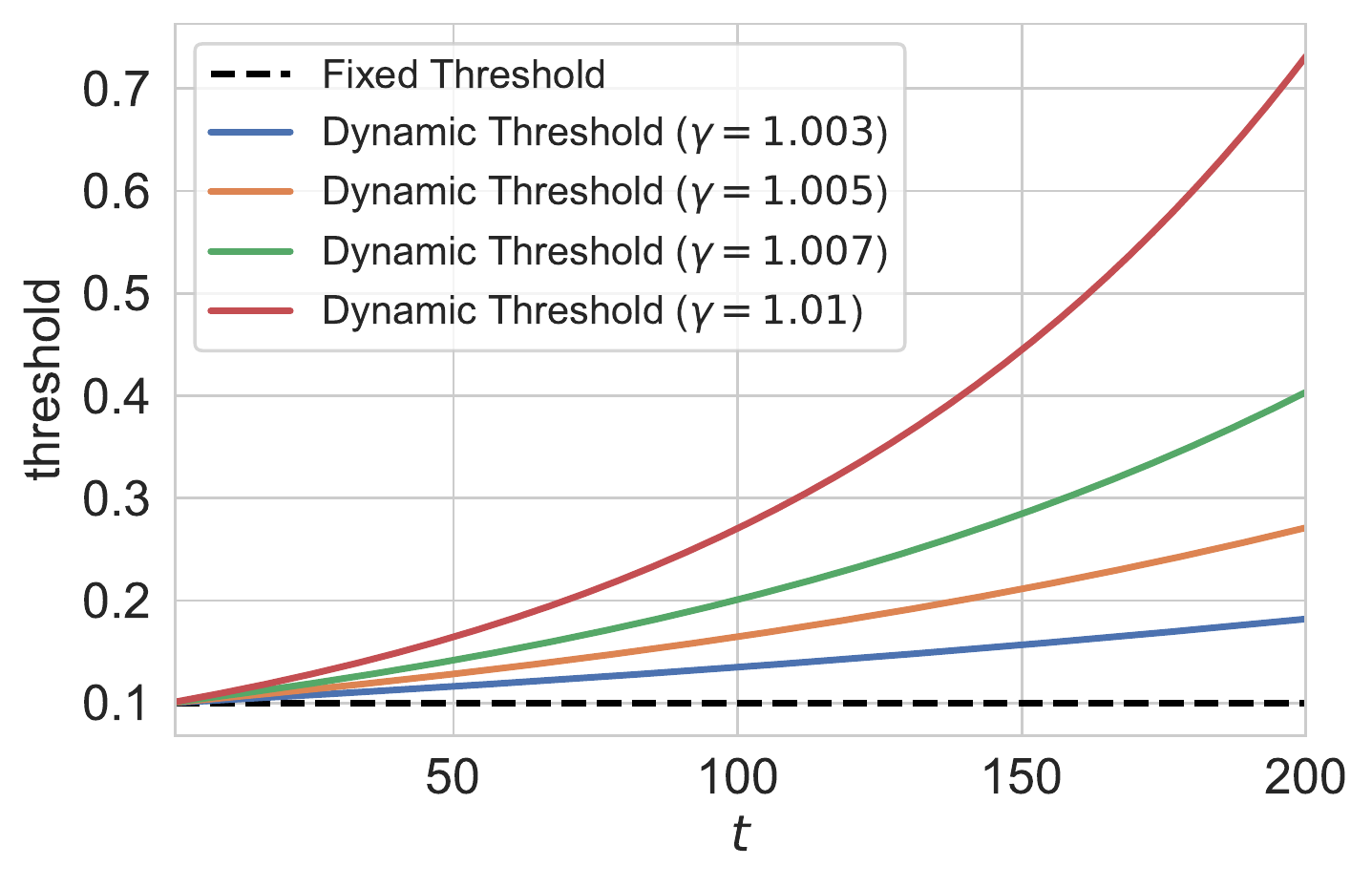}
    \caption{Exponential Scheduler}
    \end{subfigure}
    \hspace{0.1cm}
    \begin{subfigure}[b]{0.45\textwidth}
    \includegraphics[width=1\linewidth]{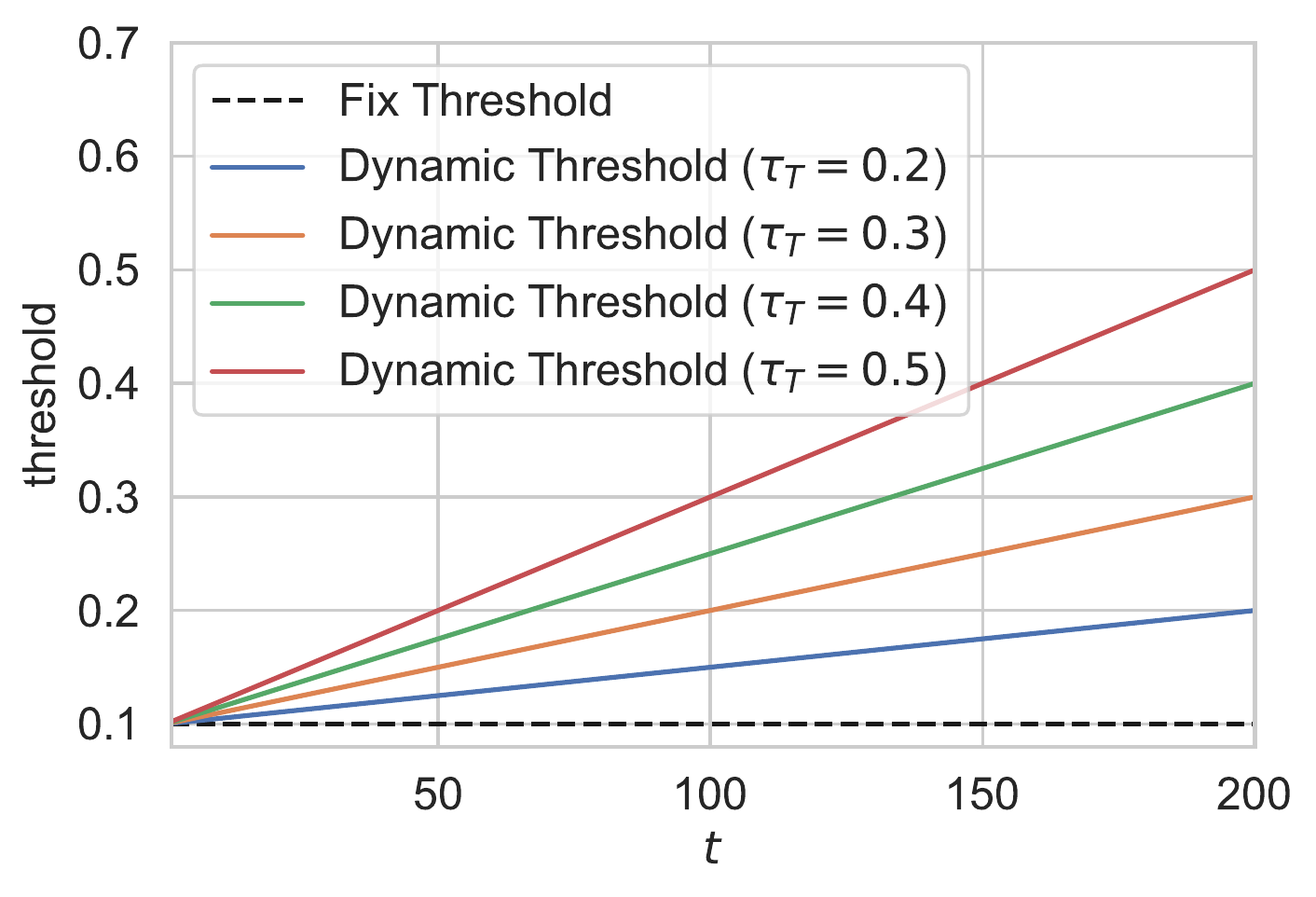}
    \caption{Linear Scheduler}
    \end{subfigure}
    \caption{Comparison of fixed threshold and dynamic threshold. Fix threshold $\tau=0.1$, exponential dynamic threshold $\tau_t=0.1\gamma^{t}$ and linear dynamic threshold $\tau_t = 0.1 + \frac{\tau_T-0.1  }{T}t$.}
    \label{fig:dynamic-threshold-demo}
\end{figure}

Figure~\ref{fig:dynamic-threshold-demo} shows a comparison of fixed threshold and the dynamic threshold $\tau_t$ with $\tau_0 = 0.1$.
We consider both exponential scheduler controlled by $\gamma$ and linear scheduler controlled by the threshold of last iteration $\tau_T$. 

We test the performance of different choice of parameters and the results are reported in Table~\ref{tab:cifar10-dynamic-threshold}. From the results, we have two observations: i) when using fixed threshold or the dynamic threshold grows too slow, performance drops in the last iterations because many noisy labels are incorrectly flagged as clean; and ii) when dynamic threshold grows too fast, the network cannot achieve best performance, because many clean labels are incorrectly flagged as noisy.

\setlength{\tabcolsep}{4pt}
\begin{table}[h]
\centering
\begin{tabular}{ c c | c | c c c  | c c c c}
\toprule
& \multirow{2}{*}{Ours ($\gamma=1.005$)} & \multirow{2}{*}{Fix} & \multicolumn{3}{ c }{Exponential} & \multicolumn{4}{ c }{Linear}\\
\cmidrule(lr){4-6} \cmidrule(lr){7-10}
 & & & 1.003 & 1.007 & 1.01 & 0.2 & 0.3 & 0.4 & 0.5 \\
 \midrule
Best & 66.33 & 66.01 & 66.27 & 63.47 & 56.81 & 65.18 & 66.09 & 61.78 & 59.41\\
Last & 64.69 & 61.37 & 63.57 & 58.93 & 35.84 & 63.40 & 65.11 & 57.84 & 55.12 \\
\bottomrule
\end{tabular}
\caption{ Test accuracy (\%) on CIFAR-10-LT with imbalance factor 100 and noise ratio 50\%.}\label{tab:cifar10-dynamic-threshold}
\end{table}

\section{Results on Clean Datasets}
Although our method is particularly designed learning with noisy labels, it is interesting to study its performance on clean but class-imbalanced datasets. In this experiment, we do not use sample re-weighting and label noise correction. We report the results in Table~\ref{tab:cifar_clean}. For fair comparison, we do not apply AugMix in this experiment.


\setlength{\tabcolsep}{7pt}
\begin{table}[!h]
\small
\centering
\begin{tabular}{ l | c c c | c c c }
\toprule
 & \multicolumn{3}{ c| }{CIFAR-10} & \multicolumn{3}{ c }{CIFAR-100} \\
\midrule
Imbalance Factor  & 10 & 50 & 100  & 10 & 50 & 100 \\
\midrule
CE  & 88.42 & 79.56 & 73.43  & 60.14 & 45.79 & 41.87  \\
LDAM  & 87.43 & 80.32 & 74.50  & 59.84 & 47.61 & 42.59  \\
LDAM-DRW  & 88.15 & 83.18 & 79.43  & 60.40 & 48.90 & 43.63  \\
cRT  & 88.26 & 79.22 & 73.61  & 60.69 & 46.67 & 42.26  \\
NCM  & 89.45 & 83.06 & 79.36  & 61.46 & 49.36 & 45.49  \\
\midrule
Prototypical Classifier  & 92.78 & 86.03 & 83.11  & 68.71 & 56.60 & 50.94  \\
\bottomrule
\end{tabular}
\caption{ Test accuracy (\%) on clean datasets with different imbalanced factor. }\label{tab:cifar_clean}
\end{table}

\end{document}